\def\year{2018}\relax
\documentclass[letterpaper]{article} 
\usepackage{aaai18}  
\usepackage{times}  
\usepackage{helvet}  
\usepackage{courier}  
\usepackage{url}  
\usepackage{graphicx}  
\usepackage{amsmath}
\usepackage{enumitem}

\begin{document}
%
\title{Learning Domain-Specific Word Embeddings from Sparse Cybersecurity Texts}

\author{Arpita Roy$^1$, Youngja Park$^2$, Shimei Pan$^1$\\
       $^1$University of Maryland, Baltimore County\\
       \{arpita2, shimei\}@umbc.edu\\
       $^2$IBM\\
       young\_park@us.ibm.com
       }
       
\maketitle
\begin{abstract}
Word embedding is a Natural Language Processing (NLP) technique that automatically maps words from a vocabulary to vectors of real numbers in an embedding space. It has been widely used in recent years to boost the performance of a variety of NLP tasks such as Named Entity Recognition, Syntactic Parsing and Sentiment Analysis.  Classic word embedding methods such as Word2Vec and GloVe work well when they are given a large text corpus. When the input texts are sparse as in many specialized domains (e.g., cybersecurity), these methods often fail to produce high-quality vectors.  In this paper, we describe a novel method to train domain-specific word embeddings from sparse texts. In addition to domain texts, our method also leverages diverse types of domain knowledge such as  domain vocabulary and semantic relations. Specifically, we first propose a general framework to encode diverse types of domain knowledge as text annotations.  Then we  develop a novel Word Annotation Embedding (WAE) algorithm to incorporate diverse types of  text annotations in word embedding.  We have evaluated our method on two cybersecurity text corpora: a  malware description corpus and a Common Vulnerability and Exposure (CVE) corpus. Our evaluation results have demonstrated the effectiveness  of our method in learning domain-specific word embeddings. 
\end{abstract}

\section{Introduction}
Word embedding is a technique in Natural Language Processing (NLP) that transforms the words in a vocabulary into dense vectors of real numbers in a continuous embedding space. While traditional NLP systems represent words as indices in a vocabulary that do not capture the semantic relationships between words, word embeddings such as those learned by neural networks explicitly encode distributional semantics in learned word vectors. Moreover, through low-dimensional matrix operations, word embeddings can be used to efficiently compute the semantics of larger text units such phrases, sentences and documents ~\cite{mikolov2013distributed}, ~\cite{le2014distributed}.  Since effective word representations play an important role in Natural Language Processing (NLP),  there is a recent surge of interests in incorporating word embeddings in a variety of NLP tasks such as Named Entity Recognition~\cite{lample2016neural},~\cite{santos2015boosting}, syntactic parsing \cite{bansal2014tailoring}, semantic relation extraction ~\cite{nguyen2015relation},~\cite{fu2014learning} and sentiment analysis ~\cite{maas2011learning}, ~\cite{tang2014learning}, to boost their performance. 

To capture the distributional semantics of words from unannotated text corpora, typical word embedding methods such as Word2Vec ~\cite{mikolov2013efficient} and GloVe~\cite{pennington2014glove} rely on the co-occurrences of a target word and  its context.  Since robust inference can be achieved  only with sufficient co-occurrences, this posts a challenge to applications where the domain texts are  sparse.   For example, traditional word embedding methods such as Word2Vec do not very perform well in highly specialized domains such as cybersecurity where important domain concepts often do not occur many times (e.g., many CVE names only occur once or twice in the entire CVE dataset). Since  word embedding is an important resource for typical NLP tasks,  a lack of  high-quality  word embeddings to capture the semantics of important domain terms and  their relations may prevent the state-of-the-art NLP techniques from being adopted in processing domain-specific texts.    


In a specific domain, in addition to text, sometimes there exist domain-specific knowledge resources created by domain experts to facilitate information processing. For example, in the medical domain, the UMLS or Unified Medical Language System defines and standardizes many health and biomedical vocabularies to enable interoperability between computer systems. In the cybersecurity domain, domain texts such as malware descriptions are often accompanied by a set of domain meta data. such as {\it malware type}. 

In this research, we proposed a general framework to leverage diverse types of existing domain knowledge to improve the quality of  word embedding when the domain texts are sparse.  First, we design a flexible  mechanism to encode diverse types of domain knowledge as text annotations. This allows us to design a unified framework to incorporate different types of domain knowledge. Previously, different word embedding algorithms have to be invented to incorporate different types of domain knowledge~\cite{ghosh2016designing,faruqui2014retrofitting}. We have also developed a Word and Annotation Embedding (AWE) algorithm that is capable of incorporating word annotations in word embeddings.   We have applied our method to two cybersecurity text corpora: a malware description corpus and a CVE  corpus. We compared the performance of our system with that of a comprehensive set of general-purpose and domain-specific word embedding models. Our evaluation results demonstrate the superiority of our method.  For example, our method outperformed the best baseline model by 22\%-57\% based on a Mean Reciprocal Rank (MRR)-based evaluation measure.

In summary, the main contributions of our work include
\begin{itemize}  
\item We present a general framework to incorporate diverse types of domain knowledge to improve the quality of domain-specific word embeddings when the input domain texts are sparse.
\begin{itemize}  
      \item We propose a flexible mechanism to encode diverse types of domain knowledge such as domain vocabulary, semantic categories and  semantic relations as text annotations.
      \item We develop a novel Word and Annotation Embedding (WAE) algorithm to incorporate text annotations in  word embeddings.  
\end{itemize}
\item We have applied the proposed method to two cybersecurity datasets.  Our method consistently and significantly outperformed a comprehensive set of baseline approaches in capturing important semantic relations between domain concepts.
\end{itemize}
\section{Related Work}
There is a rich body of work on learning general-purpose word embeddings~\cite{lecun2015deep} ,~\cite{bengio2003neural}, ~\cite{collobert2008unified}, ~\cite{mnih2009scalable}, ~\cite{mikolov2011extensions}.
Word embedding gained much popularity with the Word2Vec method ~ \cite{mikolov2013efficient}. It includes two models:  a continuous bag-of-words model (CBOW) and a skip-gram model (Skip-Gram), both learn word embeddings from large-scale unsupervised text corpora. 

Since then, many extensions to  Word2Vec have been proposed~\cite{Levy2014dependency}, ~\cite{Luong2013better}, ~\cite{yu2014improving}, ~\cite{bian2014knowledge}, ~\cite{xu2014rc}, ~\cite{faruqui2014retrofitting},  ~\cite{bollegala2016joint}, ~\cite{le2014distributed}.  For example,  Doc2Vec~\cite{le2014distributed} is an extension of word2vec which learns  vector representations for sentences and documents. Since the original work of Word2Vec uses a linear context (the words preceding and following the target word), ~\cite{Levy2014dependency} extends this by employing the syntactic contexts
derived from automatically generated dependency parse-trees. These syntactic contexts were found to capture more functional similarities, while the linear contexts in the original Skip-Gram model generate broad topical similarities.  In addition, ~\cite{Luong2013better} proposes a neural model to learn morphologically-aware word representations by combining a recursive neural network and neural language model. ~\cite{bollegala2016joint}  proposes a joint representation learning method that simultaneously predicts the co-occurrences of two words in a sentence, subject to the relational constrains given by a semantic lexicon. Finally, ~\cite{tang2014learning} learns sentiment-specific word embeddings by considering not only the syntactic 
context of words but also the sentiment of words.  

So far,  there are only very few studies focusing on learning domain-specific word embeddings.  For example, \cite{ghosh2016designing} uses information from a disease lexicon to generate disease-specific word embeddings. The main objective is to bring in-domain words close to each other in the embedding space  while pushing out-domain words away from in-domain words.  Unlike our system which can incorporate diverse types of domain knowledge, \cite{ghosh2016designing}  only concerns whether a word is in-domain or not.  To the  best of our knowledge, the method we are proposing is the most comprehensive approach for  training domain-specific word embeddings. 

\section{Datasets}
In this research, we employ two cybersecurity datasets, a  {\em malware dataset} which includes the descriptions of computer malware and a {\em CVE dataset} which includes the descriptions of common vulnerabilities and exposures in computer hardware and software. 

\subsection{The Malware Dataset}
Malware, or malicious software, is any program or file that is harmful to a computer user. Malware can perform a variety of functions including stealing, encrypting or deleting sensitive data, altering or hijacking core computing functions and monitoring users' computer activity without their permission. There are different types of malware.  For example, a malware can be a virus, a worm, a trojan horse, a spyware, a ransomware, an adware and a scareware.

\begin{figure}[h]
\centering
\includegraphics[width=8.5cm]{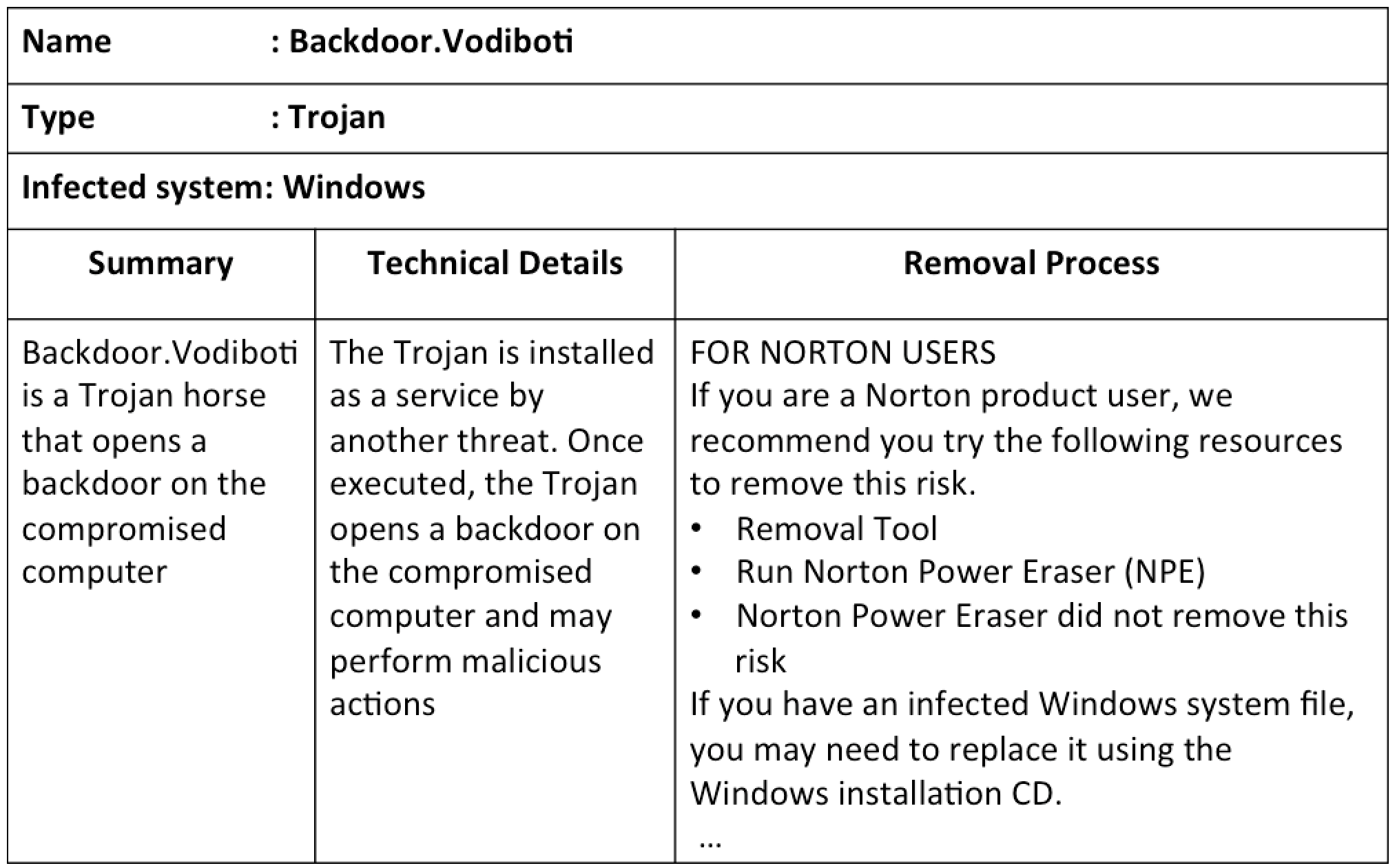}
\caption {An Example of a Malware Description }
\label{malwareemp}
\end{figure}

Our malware dataset is collected from two anti-virus companies:  Symantec\footnote{https://www.symantec.com/} and  Trend Micro\footnote{https://www.trendmicro.com/}.   The Symantec dataset contains 16,167 malware descriptions   Each  description  contains three sections:  summary,  technical details  and the removal process.  In addition, each malware description also includes a set of meta data such as  {\em malware type} and  {\em systems affected}. The Trend Micro dataset contains 424 malware descriptions.  Each  description  also includes three similar sections: overview, technical details and solution. It also includes similar metadata such as  {\em threat type} and {\em platform}, which can be roughly mapped to   {\em malware type} and  {\em systems affected} used in the Symantec description. Interestingly, different security firms often  adopt different naming conventions. As a result, the same malware may have different names due to different name conventions.  Overall, the entire malware dataset has a total of 19, 801,192 tokens and 296, 340 unique  words. Figure \ref{malwareemp} shows the Symantec description of a recently discovered malware called Backdoor.Vodiboti.

\subsection{The CVE Dataset}
In computer security, a vulnerability is a flaw or weakness in  security procedures, system design and implementation, or internal controls that could be exercised (accidentally triggered or intentionally exploited) and result in a security breach or a violation of the system's security policy. Common Vulnerabilities and Exposures (CVE) is an industry standard of common names for publicly known  security vulnerabilities and has been widely adopted by organizations to provide better coverage, easier interoperability, and enhanced security in managing cybersecurity systems. 

\begin{figure}[h]
\centering
\includegraphics[width=8.5cm]{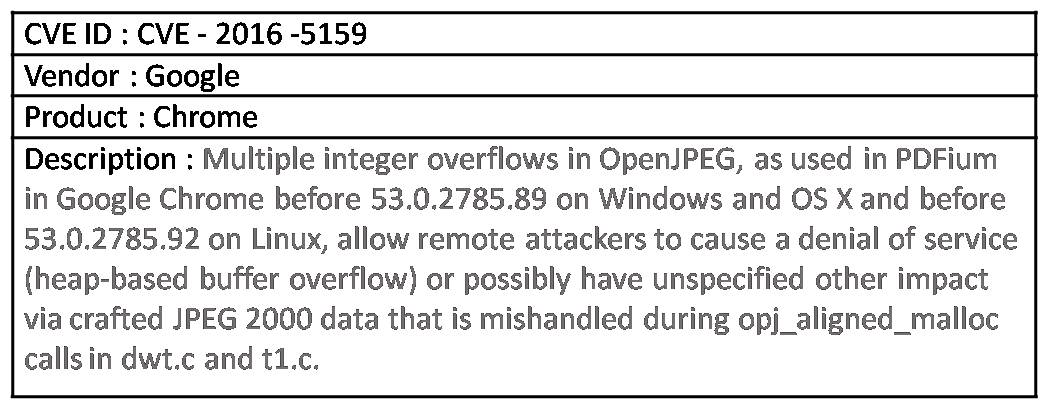}
\caption {An Example of a CVE Description }
\label{fig:CVEemp}
\end{figure}

Our CVE dataset  is collected from the  National Vulnerability Database(NVD  \footnote{https://nvd.nist.gov/}). This dataset contains detailed descriptions of  82, 871 CVEs with  a total of 11, 156, 567  word tokens and 300, 074 unique word types.  Each CVE description includes a CVE Identifier number, a  brief description of the security vulnerability or exposure, any pertinent references (i.e., vulnerability reports and advisories) and 
 additional metadata such as vendor, product and  product version. Figure \ref{fig:CVEemp} shows an example of a CVE description. 

\section{Representing Knowledge as Text Annotations }
Since there are diverse types of domain knowledge that might be useful to a specific application, instead of designing different algorithms to incorporate different kinds of domain knowledge, we develop a unified framework to incorporate different types of domain knowledge.  To facilitate this, we propose a text annotation-based mechanism to encode different types of domain knowledge such as domain vocabulary, semantic categories and semantic relations. In the following, we use an example to illustrate how to convert different types of domain knowledge into text annotations. 






\begin{figure}[h]
\centering
\includegraphics[width=8cm]{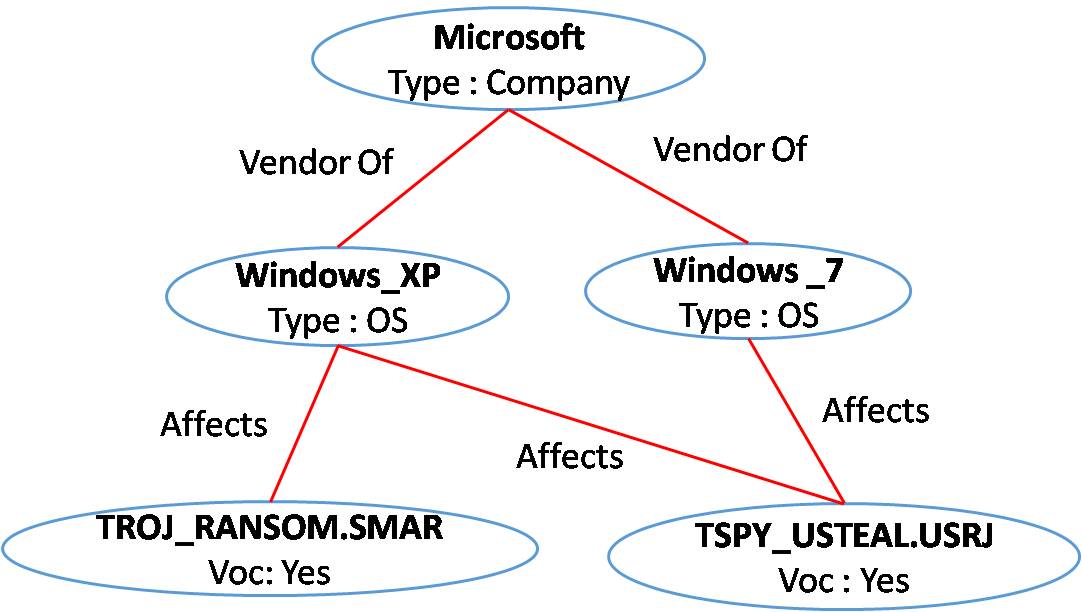}
\caption {A Graph Representing Domain Knowledge}
\label{graph}
\end{figure}

In the knowledge graph shown in Figure~ \ref{graph}, we encode four types of information (1) domain vocabulary which indicates whether a concept is in the domain or not. In this example, we assume only the malware names are in the domain vocabulary (2) semantic category which encodes the properties or metadata associated with a concept (e.g., the {\em type} of {\em Microsoft} is a {\em Company}   (3) Direct Relations which encode  direct relations between two concepts  (e.g., "Microsoft" is a "vendor of"  "Windows\_XP").  (3) Indirect Relations which encode the relations among those who share a common node in the knowledge graph. For example, Windows\_7 and Windows\_XP are connected by a common intermediate node  "Microsoft".

\begin{figure}[h]
\centering
\includegraphics[width=8.5cm]{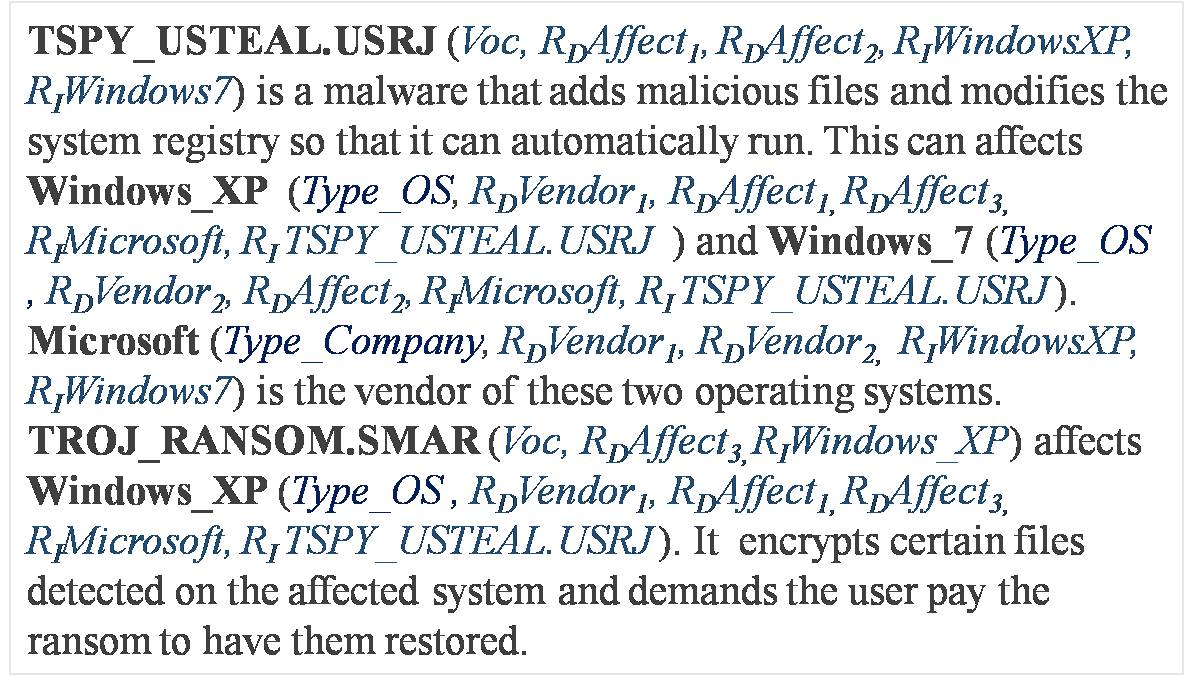}
\caption {Sample Text with Annotations}
\label{text2}
\end{figure}

To convert the graph in Figure~ \ref{graph} into text annotations, first we need to rewrite this knowledge graph into predicate-argument structures $Pred(arg_1, arg_2 ... arg_n)$. Here each predicate~{\em Pred} represents a common property or relationship shared among all the arguments. With this representation, we want to indicate that all the arguments of the same predicate should be close to each other in the word  embedding space since they either share a common property or are related by a direct or indirect relation.  With this predicate-argument representation, we can easily translate them  into text annotations. For example, the predicates will be translated to an annotation of each argument.  Table~\ref{table:PAStructure} shows the predicate-argument structures that  encode the knowledge graph in Figure~\ref{graph}. Finally,  Figure~\ref{text2} shows the  text annotated with the information in the predicate-argument structures.
\begin{table}[h!]
\begin{center}
\scriptsize
\begin{tabular}{{|l|}}
 \hline
  \textbf{Domain Vocabulary} \\
 \hline
 Voc(TROJ\_RANSOM.SMAR, TSPY\_USTEAL.USRJ)\\
 \hline\hline
 \textbf{Semantic Category}\\
 \hline 
TYPE\_OS(Windows\_XP, Windows\_7)\\
 \hline
TYPE\_Company(Microsoft) \\
 \hline \hline
 \textbf{Direct Relation}\\
 \hline 
 $R_{D}$$Vendor_{1}$=(Microsoft, Windows\_XP) \\
 \hline
 $R_{D}$$Vendor_{2}$=(Microsoft, Windows\_7) \\
 \hline
 $R_{D}$$Affect_{1}$=(TSPY\_USTEAL.USRJ, Windows\_XP)\\
 \hline
 $R_{D}$$Affect_{2}$ =(TSPY\_USTEAL.USRJ, Windows\_7) \\
 \hline
 $R_{D}$$Affect_{3}$ =(TROJ\_RANSOM.SMAR, Windows\_XP)\\  
 \hline \hline
 \textbf{Indirect Relation}\\
 \hline 
$R_{I}$$Microsoft$ =(Windows\_XP, Windows\_7)\\  
 \hline 
$R_{I}$$WindowsXP$=(Microsoft, TROJ\_RANSOM.SMAR, TSPY\_USTEAL.USRJ)\\  
 \hline 
$R_{I}$$Windows7$=(Microsoft, TSPY\_USTEAL.USRJ)\\
 \hline   
$R_{I}$$TSPY\_USTEAL.USRJ$ =(Windows\_XP, Windows\_7)\\  
 \hline 
\end{tabular}
\end{center} 
\caption{Knowledge in Predicate-Argument Structure}
\label{table:PAStructure}
\end{table}
  

In general, the nodes represented in a knowledge graph are concepts. We need to map a concept of words or terms in the text in order to generate text annotations.  Since in principle, the concept of word mapping is a one to many relation, it is possible that the same concept can be expressed by different words or phrases. For example, there are multiple ways to refer to the company "IBM" in text such as "International Business Machines", "IBM" or "Big Blue".  Since currently,  we have only implemented a simple 1 to 1 mapping,  we need to explicit enumerate different surface expressions in multiple predicate-argument structures. For example,   to indicate that IBM is the vendor of DB2, we need to produce three predicate-argument structures: $R_{D}Vendor_{3}$(IBM, DB2),  $R_{D}Vendor_{3}$(International\_Business \_Machines, DB2), $R_{D}Vendor_{3}$(Big\_Blue, DB2).  In out system, Multi-word terms  are concatenated into a single token during a pre-processing step. 

\section{Annotation and Word Embedding (AWE)}
We propose a novel AWE algorithm to learn word embeddings with text annotations. The inputs to our system are texts as well as annotations. If no annotations are available, our system is the same as a classic Word2Vec system. The output of our model includes not only a vector representation of each word but also a vector representation of each annotation in the same embedding space. For example, if {\em Type\_Trojan} is an annotation in our dataset, the algorithm will also learn a vector representation for {\em Tpye\_Trojan} based on the context of all the Trojan malware.   We have implemented two  AWE architectures. 

\begin{figure}[h]
\centering
\includegraphics[width=8cm]{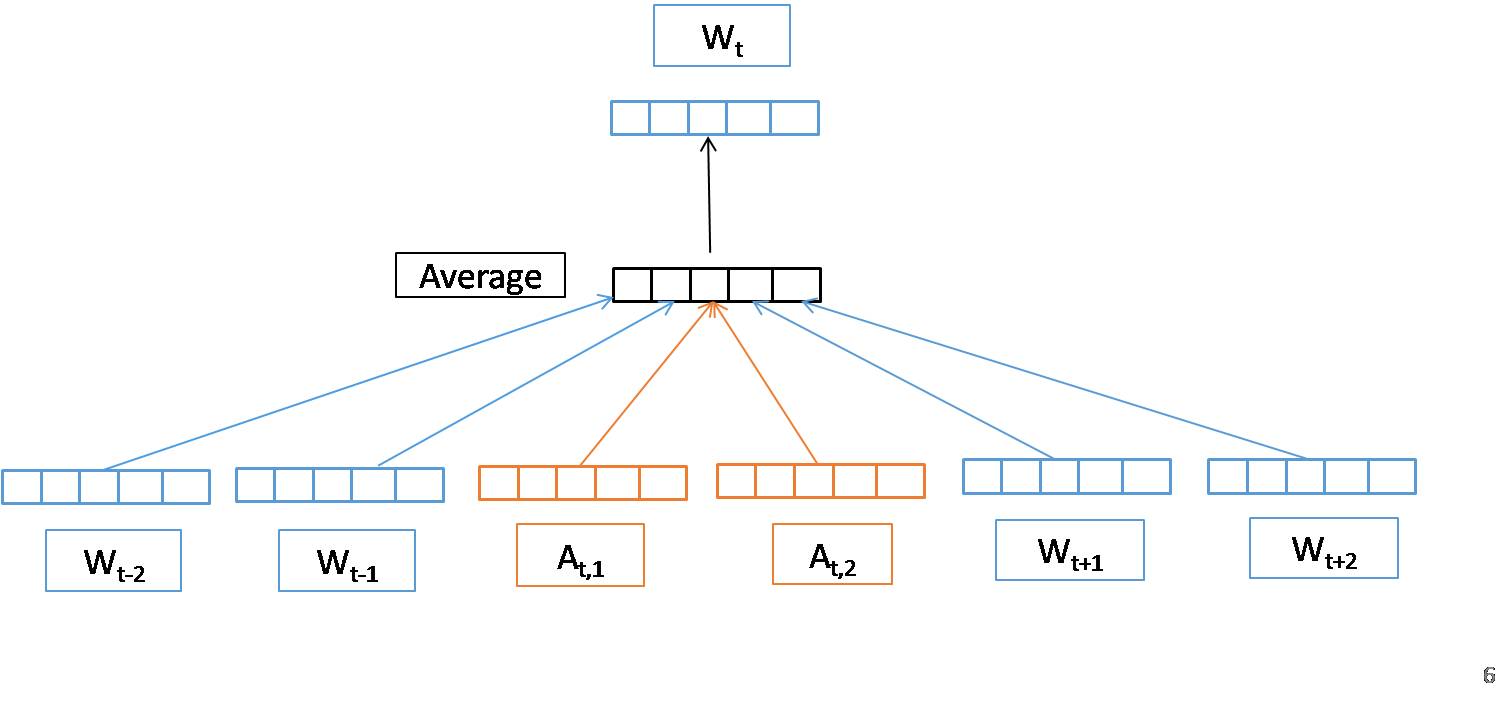}
\caption {Architecture of the AAWP Model }
\label{fig:AAWP}
\end{figure}

\textbf{ Annotation-Assisted Word Prediction (AAWP)}
With the AAWP architecture shown in Figure \ref{fig:AAWP}, the learning task is to predict a word given its annotations  plus the other words in its context. A sliding window on the input text stream is employed to generate the training samples. In each sliding window, the model tries to use the surrounding words plus the annotations of the target word as the input to predict the target word.  More formally, assume a word $W_{t}$ has a set of $M_{t}$ annotations ($A_{t,1}$, $A_{t,2}$, ... $A_{t,M_{t}}$). Given a sequence of $T$ training words $W_{1}$,$W_{2}$ ...  $W_{t-1}$, $W_{t}$ $W_{t+1}$... $W_{T}$, the objective of the AAWP model is to maximize the average log probability shown in Equation~\ref{equ:aawpobj}
\begin{equation}
\footnotesize
\frac{1}{T} \sum_{t=1}^{T}  \sum_{-C\leq j\leq C,j \neq 0} \log P(W_{t}|W_{t+j}) + \sum_{0\leq k\leq M_{t}} \log P(W_{t}|A_{t,k})
\label{equ:aawpobj}
\end{equation}

%
Where  $C$ is the size of the context window, $W_{t}$ is the target word, $W_{t+j}$ is a context word, $A_{t,k}$ is the $kth$ annotation of target word $W_{t}$. 

After training, every word is mapped to a unique vector, represented by a column in a weight matrix $Q_{w}$. The column is indexed by the position of a word in the vocabulary. Every annotation is also mapped to a unique vector, represented by a column in a weight matrix $Q_{a}$. The average of the vectors of the context words and the vectors of the annotations of the target word are then used as features to predict the target word in a context window. The prediction is done via a hierarchical softmax  classifier. The structure of the hierarchical softmax is a binary Huffman tree where short codes are assigned to frequent words. The model is trained using stochastic gradient descent where the gradient is obtained via backpropagation. After the training process converges, the weight matrix $Q_{w}$ is regarded as the learned word representations and $Q_{a}$ as learned annotation representations. 

\begin{figure}[h]
\centering
\includegraphics[width=9cm]{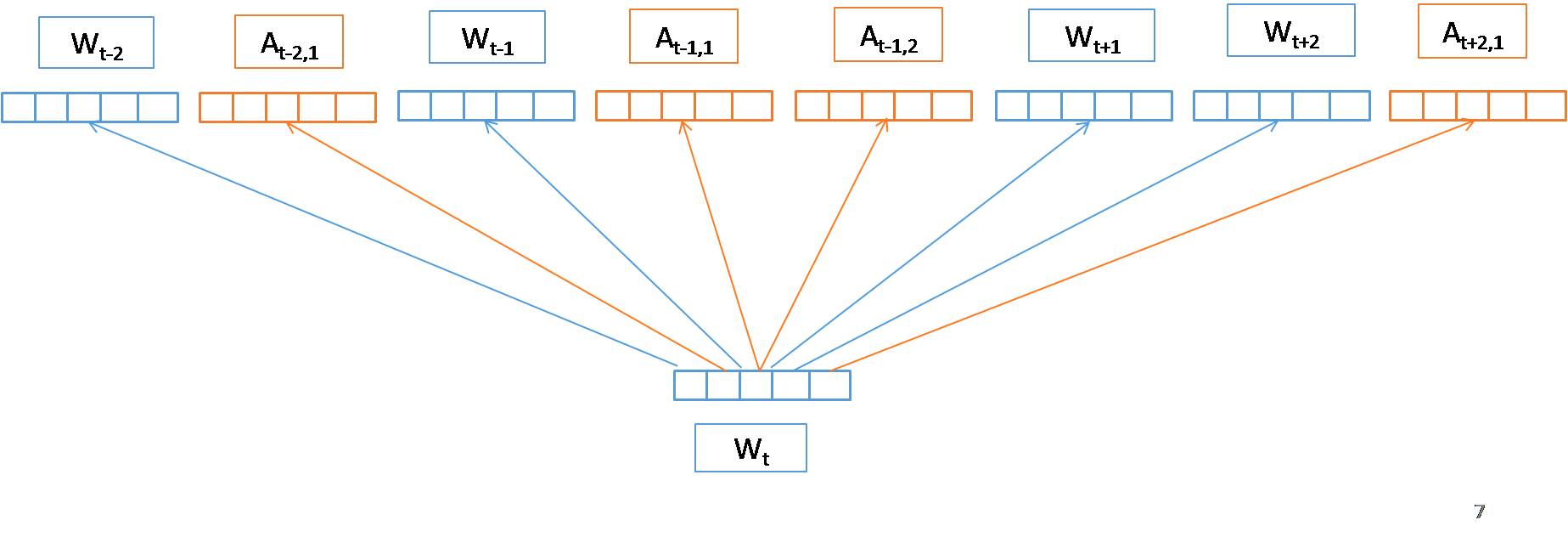}
\caption {Architecture of the JWAP model}
\label{fig:JWAP}
\end{figure}

\textbf{Joint Word and Annotation Prediction (JWAP)}
With the JWAP architecture shown in Figure \ref{fig:JWAP}, the task is to predict the context words and their annotations based on a target word. A sliding window is employed on the input text stream to generate the training samples. In each sliding window,  in addition to predicting the context words, as in typical word embedding models, if the context words have one or more annotations,  then the vector of the target word will also be used to predict the vectors of those annotations.  
  
More formally,  given a sequence of $T$ training words ($W_{1}$, $W_{2}$ ... $W_{t-1}$ $W_{t}$, $W_{t+1}$...$W_{T}$) and their annotations (($A_{1,1}, A_{1,2}...A_{1,M_{1}}),(A_{2,1}, ... A_{2,M_{2}})...(A_{T,1}, ... A_{T,M_{T}}))$, the objective of the JWAP model is to maximize the average log probability shown in Equation~\ref{equ:jwapobj}.

\begin{equation}
\scriptsize
\frac{1}{T} \sum_{t=1}^{T}  \sum_{-C\leq j\leq C,j \neq 0} \Big( \log P(W_{t+j}|W_t) + \sum_{0\leq k\leq M_{t+j}} \log P(A_{t+j,k}|W_t) \Big)
\label{equ:jwapobj}
\end{equation}

%
Where $C$ is the size of the context window,  $W_{t} $ is the target word, $W_{t+j}$ is a context word, $A_{t+j,k}$ is the $kth$ annotation of the context word $W_{t+j}$, and $M_{t+j}$ is the number of annotations associated with the context word $W_{t+j}$. The prediction is also done via hierarchical softmax.   After the training process converges, the weight matrix $Q_{w}$ is regarded as the learned word representations and $Q_{a}$ as learned annotation representations. 

\section{Evaluation} 
In this section, we describe the experiments we conducted to examine the effectiveness of the proposed method. In particular, we compare the performance of our method with that of multiple state-of-the-art baselines.  In the rest of this section, we first introduce each of the baseline models used in the evaluation, followed by a description of the evaluation methods and the results. 

\subsection{Baseline Models}
The baseline models we used includes Word2Vec ~\cite{mikolov2013efficient}, a state of the art  general-purpose word embedding system, Doc2Vec ~\cite{le2014distributed}, a variable length sentence embedding system,   Dis2Vec ~\cite{ghosh2016designing}, a domain-specific word embedding method designed to incorporate disease-related domain vocabulary in word embeddings  and a retrofitting model to incorporate semantic relations ~\cite{faruqui2014retrofitting}. 

{\bf Word2Vec:} This is one of the most widely adopted general-purpose word embedding tools.  In our experiments, we have tested different Word2Vec models with different training methods. For example, there are two separate Word2Vec models, the CBOW model and the Skip-Gram model. The CBOW model tries to maximize the log likelihood of a target word given its context words.  In contrast, the Skip-Gram model tries to maximize the log likelihood of the context words in a sliding window given a target word. In addition,  different training methods may have significant impact on their performance. For example, it has been shown that Word2Vec trained with hierarchical softmax often performed better on sparse datasets than that trained with negative sampling \cite{mikolov2013distributed}. In total, we have tested four Word2Vec models: CBOW trained with hierarchical softmax, CBOW with negative sampling, Skip-Gram with hierarchical softmax and Skip-Gram with negative sampling.
 
{\bf Doc2Vec:}  Since each malware and CVE name is associated with a text description, it is possible to represent the meaning of a malware or CVE by aggregating the meaning of all the words in  the description.  We have chosen Doc2Vec to learn a vector representation of the malware (or CVE) description.  Doc2Vec is a popular unsupervised neural network-based embedding method that learns fixed-length feature representations from variable-length texts, such as sentences, paragraphs, and documents.  In Doc2Vec every document is mapped to a document vector and every word is  mapped to a word vector. Doc2Vec also employs two different architectures:  Distributed Memory (DM) and Distributed Bag of Words (DBOW).  In DM,  the document vector and  vectors of the context words in a sliding window are aggregated together to predict the target word. In DBOW,  the document vector is used to predict the words randomly sampled from the document. 
 
{\bf Dis2Vec:} It tries to incorporate disease-related vocabulary $V$ to improve the quality of domain-specific word embeddings. Each word pair $(w,c)$ is classified into three categories:  
\begin{enumerate}
\item $D(d) = {(w, c): w \in V \cap c \in V}$ 
\item $D(\neg d) = {(w, c): w \notin V \cap c \notin V}$ 
\item  $D(d)(\neg d) = {(w, c): w \in V \oplus c \in V}$. 
\end{enumerate}

Depending on which category a word $w$ and its context word $c$ belong to, different objective functions are used. For example, when both the target and the context word are from the domain vocabulary, the objective function shown in Equation~\ref{equ:dis2vec1} is used.  The goal is to derive similar embeddings for them by maximizing the dot product of the two vectors.  
\begin{equation}
\begin{split}
l_{D_{(d)}} & = \sum\limits_{(w,c)\in D_{(d)}} 
\Big(\log \sigma(w\cdot c) \\
&+ k \cdot [P \cdot (x_{k}<\pi_{s})E_{cN\sim P_{D_{c\notin v}}}[ log \sigma (-w\cdot c_N)]\\
&+ [P \cdot (x_{k} \geq \pi_{s})E_{cN\sim P_{D_{c\in v}}} [log \sigma (-w\cdot c_N)]]\Big)
\end{split}
\label{equ:dis2vec1}
\end{equation}
Here  $x_k\sim U(0, 1), U(0,1)$ being the uniform distribution on the interval $[0,1], \alpha$ is a smoothing parameter and $\pi_s$ is a sampling parameter, $c_N$ is a negative context, $D_{c\in v}$ is the collection of (w, c) pairs for which $c \in V, D_{c\notin v}$ is the collection of (w, c) pairs for which $c \notin V$. 

For a word pair from the second category where neither the target nor the context word is from the domain vocabulary, the classic Word2Vec objective function is used (as shown in Equation~\ref{equ:dis2vec2}). 

\begin{equation}
\scriptsize
l_{D_{(\neg d)}} = \sum\limits_{(w,c)\in D_{(\neg d)}}\Big(\log \sigma(w\cdot c)+k \cdot E_{cN\sim P_D}[\log\sigma(-w\cdot c_N)]\Big)
\label{equ:dis2vec2}
\end{equation}

Finally, for word pairs from the third category where either the target word or the context word appears in  the domain vocabulary but not both, the objective function shown in Equation~\ref{equ:dis2vec3} is used. The goal is to minimize the dot product of the vectors to generate dissimilar word vectors. 

\begin{equation}
\begin{split}
l_{D_{(d)(\neg d)}}  &= \sum\limits_{(w,c)\in D_{(d)(\neg d)}} \Big(P(z<\pi_o)\log\sigma(-w\cdot c) \\ &+P(z\geq\pi_o)\log\sigma(w\cdot c)\Big)\\
\end{split}
\label{equ:dis2vec3}
\end{equation}
Where $z\sim U(0, 1), U(0,1)$ being the uniform distribution on the interval $[0,1], \alpha$ is soothing parameter and $\pi_o$ is objective selection parameter.
%

%

{\bf Retrofitting Word Vectors to Semantic Relations:} This model refines existing word embeddings using semantic relations~\cite{faruqui2014retrofitting}. This graph-based learning technique is applied as a post-processing step. Intuitively, this method encourages the new vectors to be similar
to the vectors of related words as well as to their purely distributional representations. 

Let $V$ = {$W_{1}$, . . . , $W_{n}$ be a vocabulary and Ω be an ontology that encodes semantic
relations between words in the vocabulary $V$. Let
$\Omega$ be an undirected graph (V, E) with one vertex for
each word type and edges ($W_{i}$
, $W_{j}$ ) $\in$ E $\subseteq$ $V$ x $V$ indicating a semantic relationship.  Let the matrix $\hat{Q}$ be the collection of original vector representations
$\hat{q_{i}} \in R_{d}$ for each $W_{i} \in V$ learned
using any word embedding technique, where $d$ is
the length of the word vectors. The objective is
to learn the matrix Q = ($q_{1}$,$q_{2}$...$q_{n}$) such that the columns are close to both
their counterparts in $\hat{Q}$ and adjacent vertices in $\Omega$.  Equation~\ref{equ:retrofit} shows the objective function:
\begin{equation}
\psi (Q)= \sum_{i=1}^n \Big[ \alpha_i||q_i-\hat{q_i}||^2+\sum_{(i,j)\in E} \beta_ij||q_i-q_j||^2\Big]
\label{equ:retrofit}
\end{equation}
Where $\alpha$ and $\beta$ control the relative strengths of the associations.
%


\subsection{Experiments}
To train our AWE models,  we need a text corpus plus some additional domain knowledge. Since our cybersecurity datasets include additional metadata, in our experiments,  we focus on incorporating domain metadata in word embedding. For the malware dataset, we incorporate {\em malware type} and {\em systems infected}\footnote {we mapped {\em threat type} and {\em platform} in Trend Micro to {\em malware type} and {\em systems affected} in Symantec since they are roughly the same}. For the CVE dataset, we incorporate {\em vendor} and {\em product}. All the metadata that is available to our system is also available to all the baselines as either metadata or text (some baselines can only take text as input). After we create our annotations, we train two AWE models: AAWP and JWAP.  We have experimented with different embedding vector and context window sizes and our system works the best with  embedding dimension size of 100 and context window size  of 5.   
     
To evaluate the quality of the learned embeddings by different methods, people often use the cosine similarity of the embeddings of many word pairs and its correlation (Spearman or
Pearson) with human-assessed relatedness scores \cite{schnabel2015evaluation}  However, this would require word pairs with a diverse set of similarity scores as the ground truth, which can be difficult to obtain since only experts with deep domain knowledge capable of creating such a ground truth for cybersecurity texts.  Instead, we rely on the ground truth that can be generated directly from existing data sources. Specifically, for the malware dataset, we focus on pairs of malware names that were identified as aliases by either Symantec or Trend Micro.  A total of  69 pairs of malware names were cross-referenced as aliases by at least one of the companies. For instance,
TROJ\_PSINJECT.A from Trend Micro specifies Symantec’s Trojan.Malscript as an alias.  For the CVE dataset, we created pairs of related CVE names based on whether they belong to the same CVE family and also exploit the vulnerability of the same product.  A total of 54 pairs of relevant CVEs were included in our CVE test data.

To evaluate the quality of word embeddings based on  semantically equivalent or related word pairs (e.g.,  malware aliases or relevant CVEs), we can either use the mean cosine similarity of their embeddings~\cite{mikolov2013efficient} or the Mean Reciprocal Rank (MRR)~\cite{voorhees1999trec}.  Since our focus is not to increase the absolute similarity score of the word pairs but to improve the rank of our target malware/CVE so that we can find them first in a list of most similar malware/CVEs. For example, Dis2Vec, one of the baseline models that  incorporates domain vocabulary in word embedding, brings all the in-domain words close to each other while separating the in-domain words from out-domain ones. If a simple cosine similarity based measure is used,  Dis2Vec would be considered effective since it  increases the average similarity between the word pairs in our test dataset. Since all the malware or CVE names are in the cybersecurity vocabulary, Dis2Vec brings {\em all} of them close to each other, which will not help us find the equivalent or most relevant malware/CVE first.  

Equation~\ref{equ:rank} shows the formula for computing MRR, where $M$=($m_{1}$,$m_{2}$,....$m_{T}$) is a set of $T$  malware ( or CVEs) in our domain. $P_{i}$=($m_{i,1}$ and $m_{i,2}$) is the $ith$ pair in a total of $L$ evaluation pairs. $V_{m_{k}}$ is the embedding vector for $m_{k}$,  $\bar{V}_{-m_{k}}$  is a set that includes  all  the $T$ embedding vectors excluding $V_{m_{k}}$. The function cos $(V_{x}, \bar{V}_{-x})$ produces a vector of cosine similarity scores, each is the cosine similarity of $V_{x}$ and one  the $T$ domain embedding vectors  excluding $V_{x}$ . The function $Rank (x, \bar{Y})$ returns the rank of $x$  among all the elements of $\bar{Y}$.   

\begin{equation}
\scriptsize
\begin{split}
\frac{\sum_{1\leq i \leq L}^{}Rank(cos(V_{m_{i,1}}, V_{m_{i,2}}), cos(V_{m_{i,1}}, \bar{V}_{-m_{i,2}}))} {T \times 2L} \\
+\frac{\sum_{1\leq i \leq L}^{}Rank(cos(V_{m_{i,2}}, V_{m_{i,1}}), cos(V_{m_{i,2}}, V_{-m_{i,1}}))} {T \times 2L}
\end{split}
\label{equ:rank}
\end{equation} 

\subsection{Evaluation Results}
We have tested all the embedding models on both datasets. Table \ref{table:t1} shows the results on the malware dataset. 
  
\begin{table}[h!]
\begin{center}
\begin{tabular}{{|l|l|}}
 \hline
  \textbf{Model} & \textbf{MRR} \\ 
 \hline
 \hline
 CBOW with negative sampling  & 50\% \\ 
 \hline
  CBOW with hierarchical softmax  & 44\%\\ 
 \hline
Skip-gram with negative sampling & 48\% \\
 \hline
Skip-gram with hierarchical softmax & 28\%\\ 
 \hline
Doc2Vec (DM) & 30\%  \\
 \hline
 Doc2Vec (DBOW) & 44\% \\
 \hline
  Dis2vec  & 58\% \\ 
 \hline

  Retrofitting &  28\%\\ 
  
 \hline \hline
 AAWP (new)&  41\%\\ 
 \hline
 \textbf{JAWP (new)} & \textbf{12\%}\\ 
 \hline
\end{tabular}
\end{center} 
\caption{Evaluation Results on the Malware Dataset}
\label{table:t1}
\end{table}
From the result, we can see that the JWAP model  out-performed all the other models with the highest MRR of 12\%, which represents  a 57.14\%  MRR improvement over the next best models, the  {\em Retrofitting} model and the Skip Gram model with hierarchical softmax.   In addition, among the two AWE model we proposed, the JWAP is more effective than AAWP. Here JWAP uses the target word to predict the words and annotations in its context while the AAWP model relies on the context words as well as the annotations of the target word to predict the target word. In fact, the AAWP model is a generalization of the CBOW model while the JWAP model is a generalization of the Skip-Gram model.  Previously, it was also shown that the Skip-Gram model often outperformed the CBOW model in generating general purpose word embeddings~\cite{mikolov2013efficient}.   In addition,  hierarchical softmax works better than models trained with Negative Sampling. This is also quite consistent with previous findings.  Previously, it was shown that Negative Sampling often generates better word embeddings for frequent words while hierarchical softmax works better for rare words.  Since each malware/CVE name in our dataset  occurs only once or twice,  hierarchical softmax is more effective than negative sampling in identifying semantic relations between rare words. Under the MRR measure, the vocabulary based Dis2Vec failed to improve over the general domain word embedding models.  Moreover, using document embedding to represent a domain concept (e.g., malware and CVE) does not seem to be very effective.

Table~\ref{table:t3} shows the  evaluation results on the CVE dataset.
\begin{table}[h!]
\begin{center}
\begin{tabular}{{|l|l|}}
 \hline
  \textbf{Model} & \textbf{MRR} \\ 
 \hline
 \hline
 CBOW  negative sampling  & 43\% \\ 
 \hline
 CBOW  hierarchical softmax  & 29\% \\ 
 \hline
Skip-gram negative sampling &  41\%\\ 
 \hline
Skip-gram  hierarchical softmax & 9\% \\ 
 \hline
Doc2Vec (DM) & 33\% \\
\hline
Doc2Vec (DBOW) &37 \% \\
\hline
  Dis2vec Model  & 26\% \\ 
 \hline
  Retrofitting Model & 9\% \\ 
 \hline \hline
 AAWP Model (new) & 29\% \\  
 \hline
  \textbf{JWAP model (new)} & \textbf{7\%} \\ 
 \hline
 \end{tabular}
\end{center} 
\caption{Experiment Results on CVE Dataset}
\label{table:t3}
\end{table}

The result on the CVE dataset is very similar to those on the malware dataset.  Again, the JWAP model  out-performed all the other models with the highest MRR of 7\%, which represents a 22.22\% MRR improvement over the next best systems, the retrofitting model and the Skip- Gram with hierarchical softmax model.  Among the two new models we proposed, JWAP also performed significantly better than  AAWP.   Again, the document embedding models and the Dis2Vec did not perform well.  The document vocabulary did not provide useful information to support our evaluation tasks.

\section{Discussion} 

The effectiveness of our model largely depends on the quality of the annotations. If annotations are inconsistent then our model won't be able to generate good embeddings. For example, in our malware dataset, we used {\em malware type} to generate annotations. But malware types defined by different companies are inconsistent.   For example TSPY\textunderscore SHIZ.MJSU and INFOSTEALER.SHIZ are the names of the same malware. In the Trend Micro dataset, it is categorized as a spyware while in the Symantec dataset, it is categorized as a Trojan. overall, {\em trojans} and {\em worms} are two malware types that are most inconsistent across different  companies.  Since our model tries to generate similar word embeddings for words with  the same annotations, inconsistent metadata will prevent the system from achieving this.


\section{Conclusion}
In this paper, we present a novel word embedding model for sparse domain texts.  To overcome the sparsity,  we incorporate diverse types of domain knowledge which is available in many domains. We propose a text  annotation-based framework to represent diverse types  of domain  knowledge  such as domain vocabulary, semantic categories, semantic relations and other metadata.  We have also developed a novel Word and Annotation Embedding (AWE) method which is capable of incorporating annotations in word embedding.   We have evaluated the effectiveness of our algorithm using two cybersecurity corpora: the malware dataset and the CVE dataset. We describe a series of
experiments to compare our method with  existing word embedding methods developed for both general and specific domains. Our results demonstrate that our model outperformed the next best model by  a significant margin. The improvement over the best baseline model is  22\% to 57\% MRR reduction.  The learned word embeddings  can be a  useful resource to support many downstream domain-specific NLP tasks. Currently, we are applying learned word embeddings to support Information Extraction for cybersecurity texts. 

\bibliography{reference}
\bibliographystyle{aaai}
\end{document}